\documentclass{sig-alternate-05-2015}

\usepackage{times}
\usepackage[english]{babel}

\makeatletter
\newif\if@restonecol
\makeatother

\usepackage{balance}
\usepackage{graphicx}
\usepackage{algorithm}
\usepackage{algpseudocode}

\usepackage{amsmath}
\usepackage{amssymb}
\usepackage{color}

\usepackage{multirow}
\usepackage{verbatim}
\usepackage{fixltx2e}
\usepackage{bm}
\usepackage{url}

\usepackage{eqparbox}

\usepackage{booktabs}  
\usepackage{subfig}

\usepackage{xspace}
\usepackage{microtype}

\usepackage{mathtools}
\mathtoolsset{showonlyrefs=true}

\usepackage{tikz}
\usetikzlibrary{decorations,arrows,shapes,positioning,fit,calc,plotmarks}
\usepackage{pgfplots}
\usepackage{ifpdf}

\ifpdf
\usetikzlibrary{external}
\fi

\newcommand{\scoreS}{OC$^3$\xspace}
\newcommand{\scoreI}{\textsc{BnB}\xspace}

\newcommand{\Krimp}{\textsc{Krimp}\xspace}
\newcommand{\Slim}{\textsc{Slim}\xspace}

\newcommand{\Comprex}{\textsc{CompreX}\xspace}

\newcommand{\oururl}{\url{http://eda.mmci.uni-saarland.de/bnb/}}

\newcommand{\tscore}{\mathit{score}}	

\newtheorem{definition}{Definition}[section]
\newtheorem{theorem}{Theorem}[section]

\algtext*{EndWhile}% Remove "end while" text
\algtext*{EndIf}% Remove "end if" text
\algtext*{EndFor}% Remove "end for" text

\renewcommand{\algorithmiccomment}[1]{\bgroup\hfill\scriptsize//~#1\egroup}

\algnewcommand\algorithmicinput{\textbf{Input:}}
\algnewcommand\INPUT{\item[\algorithmicinput]}

\algnewcommand\algorithmicoutput{\textbf{Output:}}
\algnewcommand\OUTPUT{\item[\algorithmicoutput]}

\renewcommand{\S}{S}	% multivariate event sequence S \in D
\newcommand{\CS}{\mathcal{S}}	% pattern set
\newlength{\tilelen} %for tikz tiles
\newcommand{\synhead}{
	\begin{tabular}{rrr c rrr c rrrr c rr c rr} 
	\toprule
	\multicolumn{3}{l}{\textbf{Data}} && \multicolumn{3}{l}{\textbf{Planted Patterns}} && \multicolumn{10}{l}{\textbf{Discovered Patterns}} \\
	\cmidrule{1-3}
	\cmidrule{5-7}
	\cmidrule{9-18}					
	&&&&&&&&&&& && \multicolumn{2}{r}{\textbf{Time (sec)}} && \multicolumn{2}{r}{\textbf{\textit{L\%}}} \\
	\cmidrule{14-15}
	\cmidrule{17-18}	
	$t(\DB)$ & $|A|$ & $|\Omega_i|$ && $|\P|$ & $||X||$ & $\support$ && $=$ & $\subset$ & $\cup$ & ? && $mean$ & $std$ && $mean$ & $std$  \\
	\midrule	
}
\newcommand{\synfoot}{\end{tabular}}

\newcommand{\support}{\mathit{support}}

\makeatletter
\renewcommand*{\@fnsymbol}[1]{\ensuremath{\ifcase#1\or \circ\or \bullet\or \ddagger\or
   \mathsection\or \mathparagraph\or \|\or **\or \dagger\dagger
   \or \ddagger\ddagger \else\@ctrerr\fi}}
\makeatother

\newcommand{\DB}{D\xspace}
\renewcommand{\P}{\mathcal{P}}

\xspaceaddexceptions{$}

\makeatletter 
\pgfdeclareshape{event}{ 
\inheritsavedanchors[from=rectangle] % this is nearly a rectangle 
\inheritanchorborder[from=rectangle] 
\inheritanchor[from=rectangle]{center} 
\inheritanchor[from=rectangle]{north} 
\inheritanchor[from=rectangle]{south} 
\inheritanchor[from=rectangle]{south east} 
\inheritanchor[from=rectangle]{south west} 
\inheritanchor[from=rectangle]{west} 
\inheritanchor[from=rectangle]{east} 
\backgroundpath{% this is new 
\southwest \pgf@xa=\pgf@x \pgf@ya=\pgf@y 
\northeast \pgf@xb=\pgf@x \pgf@yb=\pgf@y 
\pgf@yc=\pgf@ya \advance\pgf@yc by 2pt 
\pgfpathmoveto{\pgfpoint{\pgf@xa}{\pgf@yc}} 
\pgfpathlineto{\pgfpoint{\pgf@xa}{\pgf@ya}} 
\pgfpathlineto{\pgfpoint{\pgf@xb}{\pgf@ya}} 
\pgfpathlineto{\pgfpoint{\pgf@xb}{\pgf@yc}} 
} 
}
\makeatother

\tikzstyle{tile} = [rounded corners = 2pt, inner sep = 0pt, fill opacity = 0.3, anchor = south west, minimum width = 11pt, minimum height = 7pt]

\pgfdeclarelayer{background}
\pgfdeclarelayer{foreground}
\pgfsetlayers{background,main,foreground}

\tikzstyle{block} = [rounded corners, draw=blue!70, fill=white, text width=3.3cm, minimum height=4em]
\tikzstyle{bgblock} = [rounded corners, draw=blue!70, thick, fill=blue!10, text width=3.3cm, minimum height=4em]
\tikzstyle{line} = [draw, -latex', thick,blue!70]

\definecolor{yafaxiscolor}{rgb}{0.3, 0.3, 0.3}

\definecolor{yafcolor1}{rgb}{0.6, 0.016, 0.553}		%a
\definecolor{yafcolor2}{rgb}{0.7, 0.3, 0.0}			%b
\definecolor{yafcolor3}{rgb}{0.1, 0.6, 0.1}			%c
\definecolor{yafcolor4}{rgb}{0.01, 0.01, 0.01}		%d
\definecolor{yafcolor5}{rgb}{0.141, 0.345, 0.643}	%X
\definecolor{yafcolor6}{rgb}{0.4, 0.4, 0.0}			%e
\definecolor{yafcolor7}{rgb}{0.0, 0.3, 0.3}			%h
\definecolor{yafcolor8}{rgb}{0.925, 0.165, 0.224}	%Y

\newlength{\yafaxispad}
\newlength{\yaftlpad}
\newlength{\yaflabelpad}
\newlength{\yafaxiswidth}
\newlength{\yafticklen}
\makeatletter
\def\pgfplots@drawtickgridlines@INSTALLCLIP@onorientedsurf#1{}
\makeatother

\newcommand{\yafdrawaxis}[4]{
	\pgfplotstransformcoordinatex{#1}\let\xmincoord=\pgfmathresult 
	\pgfplotstransformcoordinatex{#2}\let\xmaxcoord=\pgfmathresult 
	\pgfplotstransformcoordinatey{#3}\let\ymincoord=\pgfmathresult 
	\pgfplotstransformcoordinatey{#4}\let\ymaxcoord=\pgfmathresult 
	\pgfsetlinewidth{\yafaxiswidth} 
	\pgfsetcolor{yafaxiscolor}
	\pgfpathmoveto{\pgfpointadd{\pgfpointadd{\pgfplotspointrelaxisxy{0}{0}}{\pgfqpointxy{\xmincoord}{0}}}{\pgfqpoint{-0.5\yafaxiswidth}{\yafaxispad}}}
	\pgfpathlineto{\pgfpointadd{\pgfpointadd{\pgfplotspointrelaxisxy{0}{0}}{\pgfqpointxy{\xmaxcoord}{0}}}{\pgfqpoint{0.5\yafaxiswidth}{\yafaxispad}}}
	\pgfpathmoveto{\pgfpointadd{\pgfpointadd{\pgfplotspointrelaxisxy{0}{0}}{\pgfqpointxy{0}{\ymincoord}}}{\pgfqpoint{\yafaxispad}{-0.5\yafaxiswidth}}}
	\pgfpathlineto{\pgfpointadd{\pgfpointadd{\pgfplotspointrelaxisxy{0}{0}}{\pgfqpointxy{0}{\ymaxcoord}}}{\pgfqpoint{\yafaxispad}{0.5\yafaxiswidth}}}
	\pgfusepath{stroke}
}

\pgfplotscreateplotcyclelist{yaf}{% 
{yafcolor1,mark options={scale=0.75},mark=o}, 
{yafcolor2,mark options={scale=0.75},mark=square},
{yafcolor3,mark options={scale=0.75},mark=triangle},
{yafcolor4,mark options={scale=0.75},mark=o},
{yafcolor5,mark options={scale=0.75},mark=o},
{yafcolor6,mark options={scale=0.75},mark=o},
{yafcolor7,mark options={scale=0.75},mark=o},
{yafcolor8,mark options={scale=0.75},mark=o}} 

\pgfplotsset{axis y line=left, axis x line=bottom,
	tick align=outside,
	tickwidth=\yafticklen,
	clip = false,
    x axis line style= {-, line width = 0pt, color=black!0},
    y axis line style= {-, line width = 0pt, color=black!0},
    x tick style= {line width = \yafaxiswidth, color=yafaxiscolor, yshift = \yafaxispad},
    y tick style= {line width = \yafaxiswidth, color=yafaxiscolor, xshift = \yafaxispad},
    x tick label style = {font=\scriptsize, yshift = \yaftlpad},
    y tick label style = {font=\scriptsize, xshift = \yaftlpad},
    every axis y label/.style = {at = {(ticklabel cs:0.5)}, rotate=90, anchor=center, font=\scriptsize, yshift = -\yaflabelpad},
    every axis x label/.style = {at = {(ticklabel cs:0.5)}, anchor=center, font=\scriptsize, yshift = \yaflabelpad},
    x tick label style = {font=\scriptsize, yshift = 1pt},
    grid = major,
    major grid style  = {dash pattern = on 1pt off 3 pt},
	every axis plot post/.append style= {line width=\yafaxiswidth} ,
	legend cell align = left,
	legend style = {inner sep = 1pt, cells = {font=\scriptsize}},
	legend image code/.code={%
		\draw[mark repeat=2,mark phase=2,#1] 
		plot coordinates { (0cm,0cm) (0.15cm,0cm) (0.3cm,0cm) };% 
	} 
}

\usepackage[space]{grffile}	%to plot files where filename includes whitespaces

\begin{document}

\title{Beauty and Brains:\\Detecting Anomalous Pattern Co-Occurrences}

\numberofauthors{3} 
\author{
% 1st. author
\alignauthor
Roel Bertens\\ %\titlenote{Dr.~Trovato insisted his name be first.}\\
       \affaddr{Department of Information and Computing Sciences}\\
       \affaddr{Utrecht University}\\
       \affaddr{Utrecht, The Netherlands}\\
       \email{R.Bertens@uu.nl}
% 2nd. author
\alignauthor
Jilles Vreeken\\ %\titlenote{The secretary disavows any knowledge of this author's actions.}\\
       \affaddr{Max Planck Institute for Informatics}\\
	   \affaddr{and Saarland University}\\
       \affaddr{Saarbr\"{u}cken, Germany}\\       
       \email{jilles@mpi-inf.mpg.de}
% 3rd. author
\alignauthor 
Arno Siebes\\ %\titlenote{This author is the one who did all the really hard work.}\\
       \affaddr{Department of Information and Computing Sciences}\\
       \affaddr{Utrecht University}\\
       \affaddr{Utrecht, The Netherlands}\\
       \email{A.P.J.M.Siebes@uu.nl}
}

\date{}

\maketitle
 
\begin{abstract}
	Our world is filled with both beautiful and brainy people, but how often does a Nobel Prize winner also wins a beauty pageant? Let us assume that someone who is both very beautiful and very smart is more rare than what we would expect from the combination of the number of beautiful and brainy people. Of course there will still always be some individuals that defy this stereotype; these beautiful brainy people are exactly the class of anomaly we focus on in this paper. They do not posses intrinsically rare qualities, it is the unexpected combination of factors that makes them stand out.
	
	In this paper we define the above described class of anomaly and propose a method to quickly identify them in transaction data. Further, as we take a pattern set based approach, our method readily explains why a transaction is anomalous. The effectiveness of our method is thoroughly verified with a wide range of experiments on both real world and synthetic data. 
	
\end{abstract}	

\section{Introduction}
	The recognition of anomalies provides useful application-specific insights \cite{aggarwal:13:outlier}. More specifically, the field of anomaly detection focusses on the identification of data that significantly differ from the rest of the dataset --- so different that it gives rise to the suspicion that it was generated by a different mechanism. Such an anomaly may, e.g., occur because of an error, it may be an outlier, or it may be a highly unexpected data point. It is hard, if not impossible, to distinguish between such different reasons automatically. Hence, anomalies should be inspected manually to decide whether it should, e.g., be removed, corrected, or simply remain in the data ``as is''. One should thus preferably not report an overly large list of potentially anomalous data points and, at the very least, that list should be ordered such that the most anomalous data points appear on top.
	
For transactional data anomaly detection usually boils down to pointing out those transactions that show unexpected behaviour. This unexpected behaviour can manifest itself in different ways and each detection algorithm is limited to find only those anomalies which fit the corresponding framework. For example, much work has been done to detect unexpected behaviour which can be expressed by the compressed size of a transaction given a pre-processed model \cite{smets:11:odd,akoglu:12:comprex}. That is, transactions that badly fit the norm of the data are deemed to be anomalous. Another example is to score transactions based on the number of frequent patterns that reside in it \cite{He:02:fp-outlier}. Yet another method scores transactions based on items missing from a transaction which were expected given the set of mined association rules \cite{narita:08:outlier-degree}. All these methods have their own advantages, however, none of them is able to detect an anomaly based on the presence of multiple items in a single transaction that are not expected to occur together. Therefore in this work we focus on this class of anomalies, not to improve existing methods, but to improve the field of anomaly detection by making it more comprehensive.
	
	In addition to only highlighting the transactions that show anomalous behaviour, our method describes anomalies in more detail by providing the most unlikely co-occurrence of patterns in a transaction. As an example consider a dataset containing people's drinking habits where roughly half of the people drinks Coca Cola and the other half drinks Pepsi Cola. Now each individual who drinks Coke or Pepsi is not surprising. Moreover, someone drinking both Coke and Pepsi also does not seem surprising as it can be compressed well using the methods of \cite{smets:11:odd,akoglu:12:comprex}, it contains multiple frequent patterns \cite{He:02:fp-outlier} and there is nothing missing \cite{narita:08:outlier-degree}. However, in this dataset almost everyone drinks either Pepsi or Coke, but not both. Therefore, someone drinking both Coke and Pepsi is an anomaly by definition, drinking both is unexpected. We propose to score each transaction based on the most unlikely co-occurrence between patterns and therefore our method will be able to find the described class of anomalies.
	
	For this example, the score we introduce is minus the log of the lift of the association rule Pepsi $\rightarrow$ Coke and, except the log, has been introduced before in \cite{lavrac:99:measures} as the \emph{novelty} of an association rule. The difference is that we do not score a rule, but a transaction and do so by the maximal novelty of the rules that apply to this transaction. Perhaps even more important is that we give an algorithm that does not require one to mine for all association rules with a support of 1 to find the most surprising transactions.
	
	Our new class of anomalies, their score, and the algorithm to discover these anomalies introduced in this paper is complementary to the two other well-known anomaly classes for transaction data. The first class describes transactions with a deviating length (i.e. the number of items in the transaction) and is quite trivial. The second class, for which the most work has been done \cite{smets:11:odd,akoglu:12:comprex}, describes unexpected transactions given a model of the data. To illustrate the complementarity of this latter class and our new class we show in our experiments that where current methods fail to identify our new class of anomalies, our method quickly finds them. Note, however, that our method is not intended to replace any existing methods that can discover other classes of anomalies. Rather, one should use different methods, for different classes of anomalies, that are complementary to each other.
			
The remainder of this paper is organised as follows. We first introduce notation in Section~\ref{sec:not}. In Section~\ref{sec:anomaly} we discuss the concept of anomalies in transaction data, introducing a novel class. Section~\ref{sec:how} explains how to use our score in practice. We discuss related work in Section~\ref{sec:rel}, and empirically evaluate our score in Section~\ref{sec:exp}. We round up with discussion and conclusions in Sections~\ref{sec:dis} and \ref{sec:con}, respectively.

\section{Notation} \label{sec:not}
	In this section we provide the notation used throughout the paper. We consider transaction datasets $\DB$ containing $|\DB|$ transactions. Each transaction $t$ contains a subset, of size $|t|$, of the items from the alphabet $\Omega$. Categorical data consists of $|A|$ attributes, where each attribute $A_i \in A$ has a domain $\Omega_i$, and can also be regarded as transaction data by mapping each attribute value pair to a different item. For categorical data each transaction will have the same length as each attribute should be specified. All logarithms are to base 2, and by convention 0 $\log$ 0 = 0. Further, we use $P(\cdot)$ to denote a probability density function.

\section{Anomalies in Transaction Data}	
\label{sec:anomaly}
	
		What is an Anomaly?	Anomalies are also referred to as abnormalities, discordants, deviants, or outliers in the data mining and statistics literature \cite{aggarwal:13:outlier}. As we consider transaction data we use the following definition.
		
		\begin{definition}
			A transaction is anomalous when it deviates from what we expect considering the whole dataset. 
		\end{definition}
		
		Given this definition an anomaly can manifest itself in different ways, resulting in multiple classes of anomalies for transaction data. In this section we recall 2 classes of anomalies, define 1 new class, and we show how to identify all of them by formalising appropriate anomaly scores. We want to emphasise again that the scores for different classes of anomalies are complementary to each other. Further, for ease of interpretation and computation we take the negative log-likelihood for the scores in each class.
		
	\subsection{Class 0: Unexpected Transaction Lengths}
		A transaction can be anomalous not as a result of the patterns it contains, but solely on its deviating length.
		\begin{definition}
			A class 0 anomaly is a transaction with significantly deviating transaction length.
		\end{definition}
	
		We propose an anomaly score which represents the number of bits needed to describe the transaction length given all transaction lengths in the data, i.e. for a transaction $t$ we have
		\[
			\tscore_0(t) = -\log(P(|t|)) = -\log \left( \displaystyle\frac{|\{t' \in \DB \mid |t'| = |t|\}|}{|\DB|} \right)\ .
		\]
		The intuition behind the subscript 0 for this score is that we have to take no patterns into account at all to identify these anomalies. As a result, using this score we will not be able to identify interesting co-occurrences. Further, as it is a fairly trivial score we will not further evaluate it.
	
	\subsection{Class 1: Unexpected Transactions}
		When a transaction contains none or only few frequent patterns, which do occur in (almost) all other transactions, it can be regarded to be anomalous.
		\begin{definition}
			A class 1 anomaly is a transaction that contains very little regularity.
		\end{definition}	
		The state-of-the-art in transaction anomaly detection focusses on what we call class 1 anomalies. For example, \scoreS \cite{smets:11:odd} scores transactions using a descriptive pattern set $\CS$. Transactions containing few of these patterns but mostly singletons will get a higher score. That is, because such a transaction cannot be explained well by the pattern set that is descriptive for the data. 		
		We generalise this idea by defining a score based on the probability of a transaction. More formally, $\tscore_1$ scores each transaction based on the number of bits needed to describe it, i.e. for a transaction $t$ we have		
		\[
			\tscore_1(t) = -\log P(t)  \quad .
		\]
		For compression based methods such as \scoreS this score is defined by the compressed length of the transaction given the model of the data. However, any method that can assign a probability to a transaction based on the whole data can be used here. Note that, as transactions are scored as a whole, this approach will unlikely detect unexpected co-occurrences of patterns. For example, using \scoreS, all patterns that describe a transaction will contribute to a its score independently. As much work has been done to detect these anomalies we will not further evaluate their identification, but focus on the next class of anomalies.
				
	\subsection{Class 2: Unexpected Co-occurrences}		
		The focus of this paper lies on identifying unexpected co-occurrences of patterns. To the best of our knowledge we are the first to address these class 2 anomalies. Before we explain how to identify them, we start with a definition.
		 
		\begin{definition}	\label{def:unexano}	
			A transaction contains a class 2 anomaly when it contains two patterns which occur much less together in the data than what could be expected from their individual supports.
		\end{definition}
				
		As this definition is somehow the opposite of that of a pattern, which is formed when two smaller patterns occur together more frequently than expected, we can also use the terms negative pattern or negative interaction pattern for a class 2 anomaly. These negative patterns cannot be identified using currently available methods as these do not consider negative interaction patterns. An example of a class 2 anomaly is someone drinking both Coke and Pepsi when everybody else only drinks either Pepsi or Coke. In other words, a very rare co-occurrence of the two frequent patterns Coke and Pepsi.
				
		Now to identify anomalous transactions based on class 2 anomalies we would like to score a transaction higher the more unexpected a co-occurrence of patterns it contains is. That is, we propose to rank a transaction based on its most unexpected co-occurrence. Intuitively this means that for each transaction we compute the number of bits we need to explain the most unlikely co-occurrence given a pattern set $\CS$ and the data. For a transaction $t$ we thus have \[
			\tscore_2(t) =  \hspace{-1em}\max\limits_{\{X,Y \in \CS \mid X,Y \subseteq t\}} \hspace{-1em}{ - \log P(XY) + \log\big(P(X) \times P(Y)\big) } \; .
\]
		In the remainder of this paper we refer to $\tscore_2$ as the \scoreI (Beauty and Brains) score. We compute $P(X)$ as $X$'s support or relative frequency in the data. A compression based approach similar to \scoreS to compute $P(X)$ based on its relative usage makes no sense here as we are looking for unexpected co-occurrences and are not trying to describe the entire transaction.
		
		Given a \scoreI score for a transaction we can readily explain its anomalousness as we know which co-occurrence of patterns is responsible for the score. Therefore our method has the nice property of producing very interpretable rankings.
			
		Our score is related to the concept of lift~\cite{shapiro:91:lev} used in the context of association rules. In our setting we use it to describe the difference between two patterns appearing together in a transaction and what would be expected if they were statistically independent. Therefore, the higher our score the more unexpected the pattern co-occurrence. 
				
		Scores that are constructed to identify class 1 anomalies are not able to detect these class 2 anomalies as they look at all patterns \emph{independently}. For example, \scoreS \cite{smets:11:odd} and \Comprex \cite{akoglu:12:comprex} will not give a class 2 anomaly a higher score as both individual patterns are frequent and will add little to the anomaly score. Similarly, the frequent pattern based method from He et al.~\cite{He:02:fp-outlier} and the method from Narita et al.~\cite{narita:08:outlier-degree} have no means to give higher scores to class 2 anomalies. As a result, methods for identifying class 1 anomalies do not identify unexpected co-occurrences, while these actually do indicate anomalous behaviour.

		\subsubsection*{Which patterns to consider}\label{sec:compl}
			Given the relation between our score and the lift of association rules, a straightforward way to find high scoring transactions may seem to simply mine for high-lift association rules. However, to maximize the score the individual patterns $X$ and $Y$ should have a support as high as possible while $XY$ should have a support as low as possible.	That is we should mine for all rules including those with a support of 1 to ensure that we don't miss the most interesting transactions.
			
			Clearly this becomes infeasible for all but the smallest data sets quickly. Not only because discovering all these rules will take an inordinate amount of time, but also since the post-processing of all these rules necessary to identify the most surprising transactions becomes an even more daunting task.	
			
			The alternative we take is by starting from a set of patterns $\CS$. We compute the score of each pair of patterns from $\CS$ and identify those transaction in which pairs with a (very) high score occur.
			
			Clearly, not just any pattern set will do as want to find the highest scoring transactions. The set of all frequent patterns $\mathcal{F}$ will be very descriptive, yet far too large to be able to consider the interactions between each pair of patterns. For all but the smallest of datasets this will quickly yield infeasibly large pattern sets. That is, worst case we need to consider each co-occurrence of patterns for each transaction, thus leading to a computational complexity of
			\[
				O( |\DB| \times \mathcal{|F|} \times \mathcal{|F|} ) \quad .
			\] 
			Choosing a higher minimum support will yield smaller pattern sets but as a result we might miss important patterns. We could use condensed representations such as closed \cite{pasquier:99:discovering} or non-derivable \cite{calders:07:ndidami} frequent patterns to remove as much redundancy as possible, however these sets will still be too large. By sampling \cite{hasan:09:musk} patterns we can attain small sets of patterns, however, the choice of the size of the sample determines which anomalies one will (likely) find. A set that is too small might miss some important patterns, but a set that is too large probably contains redundancy and becomes a bottleneck in our approach as we need to look at each combination of patterns in the set. Since it is not straightforward to choose the right size for the required pattern set, we choose to use \Krimp \cite{vreeken:11:krimp} or \Slim \cite{smets:12:slim} to automatically find small descriptive pattern sets that describe the data well without containing noise or redundancy. Using these pattern sets it will hold that $|\CS| \ll |\mathcal{F}|$. We thus dramatically reduce the complexity, making the \scoreI score practically feasible as we will show in our experiments in Section~\ref{sec:exp}. Using such a vastly smaller set induces, of course, the risk that we miss anomalies. However in other research we have seen that the pattern sets chosen by \Krimp and \Slim are highly characteristic for the data. The experiments in Section~\ref{sec:exp} bear out that this is also the case here: all anomalies we inject in synthetic data are discovered using these small sets only.
			
			Next to computational complexity, another advantage of small descriptive pattern sets is that they are more easily interpretable, which is convenient when explaining the identified anomalies.

		\begin{figure}[t] 
		\centering					
			\includegraphics[width=8cm]{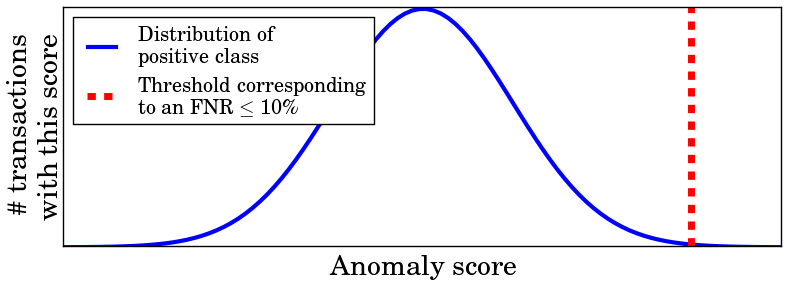}				
			\caption{\textbf{Example of setting a threshold using Cantelli's inequality.} Based on the positive class we compute a threshold corresponding to a false-negative rate of 10\%. }
			\label{fig:cant}
		\end{figure}

\vspace{1em}
		\section{How to use our scores}\label{sec:how}

	In the process of explorative data mining, one has to consider that all 3 classes of anomalies we identify give different insight. In practice, one should instantiate all 3 scores and investigate the top-ranked anomalies for each class. Here, our focus is of course on class 2 anomalies.
	
	To determine which of the \scoreI top-ranked transactions to investigate, as well as to verify the significance of the scores, we propose the following two methods based on the Bootstrap. Recall that bootstrap methods consider the given data as a sample, and generate a number of pseudo-samples from it; for each pseudo-sample calculate the statistic of interest, and use the distribution of this statistic across pseudo-samples to infer the distribution of the original sample statistic \cite{cameron:08:bootstrap}.		
	
	\subsection{Significance test} \label{sec:sig}
		For a synthetic dataset it is easy to test the significance of anomaly scores, as we can generate data with and without anomalies for which the resulting scores must clearly differ. For real world data this is unfortunately not the case as we do not know which and how much (negative) patterns the data comprises. Nevertheless, to give a measure of significance we use the following bootstrap approach. We randomly sample transactions from our original dataset (with replacement) to retrieve an equally sized new dataset. We repeat this a thousand times and save the highest anomaly score for each dataset. Then we repeat this process, but we first remove the transaction with the highest \scoreI score from the sample set. That is, the top-ranked anomaly is definitely not present in the bootstrap samples of the second kind and may or may not be present in the bootstrap samples of the first kind. The bigger the difference between the distributions of scores with and without the top-ranked transaction, the more significant the top-ranked anomaly. 			

	\subsection{Which transactions to investigate} \label{sec:inves}
		Choosing the right parameter value is never easy in explorative data mining. As the \scoreI score produces a ranking on al transactions, where higher scores indicate a higher chance on being anomalous, it does not need any parameters. To determine which transactions to investigate based on this ranking we employ Cantelli's inequality to identify the transactions that significantly differ from the norm.

		\begin{theorem}
			\textbf{Cantelli's inequality} \cite{grimmett:01:prob}. Let $X$ be a random variable with expectation $\mu_X$ and standard deviation $\sigma_X$. Then for any $k \in \mathcal{R^+}$,
			\[
				P(X - \mu_X \ge k\sigma_X) \le \frac{1}{1+k^2}	\quad .
			\]	
		\end{theorem}
		Smets and Vreeken \cite{smets:11:odd} proposed a well-founded way to determine threshold values to distinguish between `normal' and anomalous transactions. The positive class comprises anomaly scores for `normal' transactions and based on the distribution of these scores we can choose a threshold by choosing an upper bound on the false-negative rate (FNR). For example, if we choose a confidence level of 10\%, Cantelli's inequality tells us that this corresponds to a threshold $\theta$ at 3 standard deviations from the mean, given by $\theta = \mu + k \sigma$. This means that the chance on a future transaction with an anomaly score above the threshold is less than 10\%, see Figure~\ref{fig:cant}.

		To compute these thresholds we need the distribution of the positive class, i.e. the anomaly scores for all `normal' transactions. Unfortunately, we only have the one dataset available which can contain both transactions from the positive and negative (actual anomalies) class. As by definition the number of anomalies must be relatively small we use the entire dataset to estimate the distribution of the positive class again using a bootstrap approach. That is, we generate bootstrap datasets by randomly sampling transactions (with replacement) from the original dataset. We then use all anomaly scores from all bootstrap datasets to estimate the distribution.

\section{Related Work} \label{sec:rel}
	In this paper we study anomaly detection in binary transaction data. As anomalies are referred to in many different ways, mostly with slightly different definitions, we refer to \cite{markou:03:novelty} and \cite{aggarwal:13:outlier} for in-depth overviews on this field of research. In general, most anomaly detection methods rely on distances. Here we focus on discrete data, nominal attributes, for which meaningful distance measures are typically not available. 
	
	Of the methods that are applicable on transaction data, that of Smets and Vreeken \cite{smets:11:odd} is perhaps the most relevant. They propose to identify anomalies as those transactions that cannot be described well by the model of the data, where as models they use small descriptive pattern sets. Their method \scoreS works very well for one-class classification, however it is not able to identify unexpected co-occurrences in the data. Akoglu et al. \cite{akoglu:12:comprex} proposed \Comprex which takes a similar approach in that they also rank transactions based on their encoded length. The difference is that they do not use a single code table to describe the data, but a code table for each partition of correlated features. Although this method achieves very good results it is only suitable for categorical data and not for transaction data in general. 
	Note that, following our generalised anomaly score for class 1 anomalies, any method that provides a probability for a transaction can be used. Examples based on pattern sets are those of Wang and Parthasarathy \cite{wang:06:summaxent} and Mampaey et al.~ \cite{mampaey:12:mtv}. 
	
	He et al. \cite{He:02:fp-outlier} rank transactions based on the number of frequent patterns they contain given only the top-$k$ frequent patterns, and Narita et al. \cite{narita:08:outlier-degree} rank transactions based on confidence of association rules but need a minimum confidence level as parameter. All these methods have no means to identify class 2 anomalies.	
	
	In the Introduction we already mentioned the relation between our score and \emph{novelty} as introduced in \cite{lavrac:99:measures}. As stated there, the difference is that we score transactions rather than rules and we give an algorithm to quickly discover the highest scoring transactions. Our notion of anomaly is also related to the conditional anomalies introduced in \cite{song:2007:conditional}. In our running example, Pepsi could be seen as the context that makes a purchase of Coke unexpected in their terminology. The difference is that we do not expect the user to define such contexts, they are discovered automatically. Moreover, we use a small set of patterns to discover all the class-2 anomalies rather than probabilistic models on context and other attributes.
		
	To compute the \scoreI score we need a small pattern set that contains the key patterns of the data. In general, we can use the result of standard frequent pattern mining~\cite{agrawal:96:fast,pasquier:99:discovering} although this incurs a high computational cost. Instead, we can resort to pattern sampling techniques~\cite{hasan:09:musk,boley:11:direct}, yet then we have to choose the number of patterns to be sampled. Instead, Siebes et al.~ \cite{siebes:06:item} proposed to mine such pattern sets by the Minimum Description Length principle \cite{grunwald:07:book}. That is, they identify the best set of patterns as the set of patterns that together most succinctly describe the data. By definition this set is not redundant and does not contain noise.  \Krimp \cite{vreeken:11:krimp} and \Slim \cite{smets:12:slim} are two deterministic algorithms that heuristically optimise this score. Other pattern set mining techniques, especially those that mine patterns characteristic for the data such as~\cite{mampaey:12:mtv,geerts:04:tiling,wang:06:summaxent}, are also meaningful choices to be used with \scoreI.

\section{Experiments} \label{sec:exp}
	In this section we evaluate the power of the \scoreI score to identify class 2 anomalies. Firstly, we show how we generated synthetic data needed for some of the experiments. Secondly, we provide a baseline comparison where we show that the size of the input set of patterns is of great importance. Next we show the performance of \scoreI on synthetic data and show its statistical power. Lastly, we show some nice results of class 2 anomalies on a wide variety of real world datasets. 
	
	We implemented our algorithms in C++ and generated our synthetic data using Python. Our code, both to compute anomaly scores and to generate synthetic data, is available for research purposes.\!\footnote{\oururl} All experiments were conducted on a 2.6 GHz system with 64 GB of memory. 
	
	\subsection{Generating Synthetic Data}
		Here we describe how we generated both transaction and categorical synthetic data.
			
		\subsubsection*{Transaction Data} \label{sec:gentra}
			To generate synthetic transaction datasets we first choose the number of transactions $|\DB|$ and the size of the alphabet $|\Omega|$. Further, we generate a set of random patterns $\mathcal{P}$ and for each pattern in this set we choose a random support in the range [5-10\%] and a random size from 3 to 6 items. In addition we similarly generate 2 more patterns with a support of 20\%, which we call the anomaly generators. Then we build our dataset by first adding the 2 anomaly generators for which we make sure that they only occur together in the same transaction once; that is the anomaly. Thereafter, we do the following for each transaction. With the probability corresponding to its support each pattern from $\mathcal{P}$ is added to the transaction as long as it does not interfere with the anomaly. In addition, each singleton from $\Omega$ is added to each transaction similarly with a probability of 10\%.				
			
		\subsubsection*{Categorical Data}\label{sec:gencat}
			To generate categorical data we take a similar approach. Firstly, we choose the number of transactions $|\DB|$, the number of attributes $|A|$ and the alphabet size per attributes $|\Omega_i|$. We generate random patterns with the same settings as for transaction data and again first add the anomaly generators to the dataset. We then try to add the other patterns as long as they fit and do not interfere with the anomaly. Then we fill the unspecified attributes for each transaction with random singletons.

	\subsection{Baseline Comparison}
\begin{figure}[t] 
			\centering					
				\includegraphics[width=8cm]{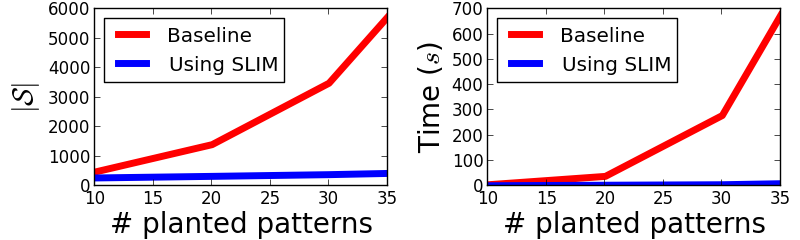}				
				\caption{\textbf{\Slim pattern set versus closed frequent patterns (baseline).} We computed the \scoreI scores on 5\,000 transactions using both input sets with a minimum support of 1 for \Slim and at 5\% for the baseline. Using both input sets $\mathcal{S}$ the anomaly is always ranked first, but for the baseline both $|\mathcal{S}|$ and the runtime quickly explode when we increase the number of planted patterns in the data. The runtimes include the time needed to compute the used input set. }
				\label{fig:baseline}
		\end{figure}

		In the following sections we will show that \scoreI gives very reliable scores, but first we show its efficiency here. To emphasise the necessity for using small pattern sets as input for our \scoreI score we compare the use of \Slim \cite{smets:12:slim} pattern sets with a minimum support of 1 against the use of all closed frequent patterns with a minimum support at 5\%. We generated random transaction data with $|\DB|$ = 5\,000, $|\Omega|$ = 50 and we ranged $|\mathcal{P}|$ from 10 to 35 patterns. We then ran our method on both input sets keeping track of the runtimes and the size of the input set $\CS$ for which we have to consider all $|\CS| \times |\CS|$ possible combinations. Both approaches always rank the anomaly highest, therefore further we can focus on the runtime and the number of patterns that were considered. Note that, the runtimes include the time needed to compute the used pattern sets, however, these are negligible as the exponential growth in runtime for the baseline approach is caused by the exponential growth of $\CS$. Figure~\ref{fig:baseline} shows the results which are averages over 5 runs per setting. For higher minimum support thresholds the baseline approach starts missing important patterns and it cannot identify the anomaly. Other settings for generating synthetic data lead to a similar figure. Since using a \Slim pattern set as input set for \scoreI we attain similar results compared to the baseline approach, that is we correctly identify the planted anomaly, in the remainder of this paper we always use the \Slim pattern set to compute the \scoreI score.

	\subsection{Performance on Synthetic Transaction Data}
		The goal of this experiment is twofold. Firstly, we show that our method is able to identify class 2 anomalies in transaction data. Secondly, we justify the definition of the different classes of anomalies as we show that the class 2 anomalies are not identified by the state-of-the-art class 1 anomaly detector, which is \scoreS \cite{smets:11:odd}. We emphasise again that as a result both scores should not be further compared as they are complementary to each other.
		
		We generated random datasets as described in Section~\ref{sec:gentra} with various settings. The results in Table~\ref{tab:syn-tra} show that \scoreI always ranks the anomaly highest and that \scoreS does not identify them.
		
		\begin{table}[t] 		
			\centering
			\caption{\textbf{The performance of \scoreI on transaction data.} The number of generated transactions is represented by $|\DB|$, the alphabet size by $|\Omega|$, and the number of synthetic patterns by $|\mathcal{P}|$. All experiments were performed 10 times and the average ranks and runtimes (in seconds) are reported.}
			\label{tab:syn-tra}				
			\begin{tabular}{cccc@{}ccc@{}cc} 
				\toprule
				\multicolumn{3}{l}{\textbf{Generated Data}} && \multicolumn{2}{l}{\textbf{\scoreI}} && \multicolumn{2}{l}{\textbf{\scoreS}} \\
				\cmidrule{1-3}
				\cmidrule{5-6}
				\cmidrule{8-9}
				
				$|\DB|$ & 					
				$|\Omega|$ &
				$|\mathcal{P}|$ &
				& 
				rank & 
				time ($s$)&
				&
				rank &
				time ($s$)\\
				
				\midrule	
				5\,000	&	50	&	100	&&	1 & 4	&& 2\,420 & 1 \\
				5\,000	&	100	&	100	&&	1 & 6	&& 2\,757 & 2  \\
				5\,000	&	100	&	200	&&	1 & 18	&& 2\,433 & 4 \\				
				10\,000	&	100	&	100	&&	1 & 11	&& 5\,464 & 4 \\
				20\,000	&	50	&	100	&&	1 & 18	&& 8\,281 & 4 \\				
				\bottomrule		
			\end{tabular}	
		\end{table}
				
	\subsection{Performance on Synthetic Categorical Data}
		Knowing that \scoreI correctly identifies class 2 anomalies for transaction data, here we compared it to the state-of-the-art on categorical data, which is \Comprex \cite{akoglu:12:comprex}. Again we note that we only compare these methods to show that class 2 anomalies are different from class 1 anomalies and that these two methods thus should be used complementary to each other.
		
		We generated random datasets as described in Section~\ref{sec:gencat} with various settings. The results in Table~\ref{tab:syn-cat} show that \scoreI always ranks the anomalous transaction first and \Comprex is not able to identify it (gives it a much lower rank).		
		\begin{table}[t] 		
			\centering
			\caption{\textbf{The performance of \scoreI on categorical data.} Each dataset contains 5\,000 transactions over $|A|$ attributes, each with an alphabet size of $|\Omega_i|$. The number of synthetic patterns is refered to by $|\mathcal{P}|$. All experiments were performed 10 times. We report the average ranks and runtimes (in seconds).}
			\label{tab:syn-cat}				
			\begin{tabular}{cccc@{}ccc@{}cc} 
				\toprule
				\multicolumn{3}{l}{\textbf{Generated Data}} && \multicolumn{2}{l}{\textbf{\scoreI}} && \multicolumn{2}{l}{\textbf{\Comprex}} \\
				\cmidrule{1-3}
				\cmidrule{5-6}
				\cmidrule{8-9}
								
				$|A|$ &
				$|\Omega_i|$ &
				$|\mathcal{P}|$ &
				& 
				rank & 
				time ($s$)& 
				&
				rank & 
				time ($s$) \\ 
				
				\midrule	
				20	&	5	&	100	&&	1	&	1	&& 3\,119	& 221		\\
				50	&	5	&	100	&&	1	&	6	&& 2\,028 	& 885		\\
				100	&	5	&	100	&&	1	&	29	&& 3\,121 	& 5\,477	\\
				20	&	10	&	100	&&	1	&	1	&& 2\,244 	& 429		\\				
				50	&	10	&	200	&&	1	&	5	&& 2\,714 	& 1\,978	\\				
				\bottomrule		
			\end{tabular}	
		\end{table}

	\subsection{Statistical Power}
\begin{figure}[h] 
			\centering					
				\includegraphics[width=8cm]{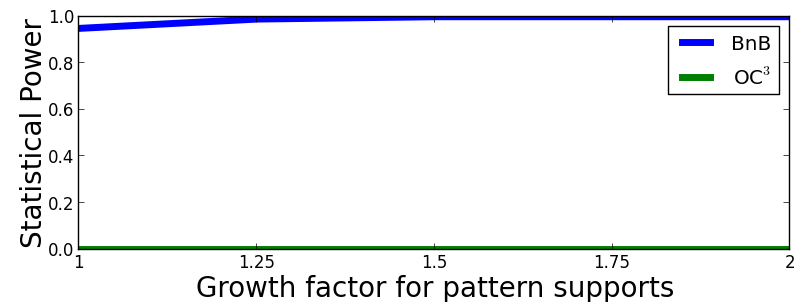}				
				\caption{[Higher is better] \textbf{Statistical power of \scoreI.} Whereas \scoreS does not identify any class 2 anomalies, \scoreI does perfectly with large enough supports. We observe the same behaviour for categorical data comparing \scoreI and \Comprex (not shown in this plot). The growth factor on the x-axis describes the increase of the pattern supports in the data.}
				\label{fig:statpow}
		\end{figure}

		Our aim here is to examine the power of the \scoreI score for identifying class 2 anomalies. For this purpose, we perform statistical tests using synthetic data. To this end, the null hypothesis is that the data contains no class 2 anomalies. To determine the cutoff for testing the null hypothesis, we first generate 100 transaction datasets without the single co-occurrence between the 2 anomaly generators, whereafter we generate another 100 datasets with this co-occurrence included. For all datasets we choose $|\DB|$ = 5\,000, $|\Omega|$ = 25 items and $|\mathcal{P}|$ = 100. Next, we report the highest \scoreI score for all 100 datasets without anomaly. Subsequently, we set the cutoff according to the significance level $\alpha = 0.05$. The power of the \scoreI score is the proportion of the highest scores from the 100 datasets with anomaly that exceed the cutoff. Note that, we only look at the highest score for each dataset as we know that this must be the anomaly for the datasets containing it. We show the results in Figure~\ref{fig:statpow} while varying the the range from which we randomly choose the supports for the patterns in $\mathcal{P}$ from [4-8\%] to [8-16\%] and the support for the anomaly generators from 16\% to 32\%. In Figure~\ref{fig:statpow} we label these linearly growing supports with their growth factor from 1 to 2. With other settings to generate the data we observe the same trend. Again, only to emphasise that methods to identify class 1 anomalies are not suitable to discover class 2 anomalies, in Figure~\ref{fig:statpow} we also plotted the statistical power of \scoreS regarding class 2 anomalies. As \Comprex is not applicable to transaction data we performed a similar experiment on categorical data. This resulted in a similar plot with \scoreI at the top and \Comprex at the bottom.
				
\begin{figure}[t] 
			\centering					
				\includegraphics[width=8cm]{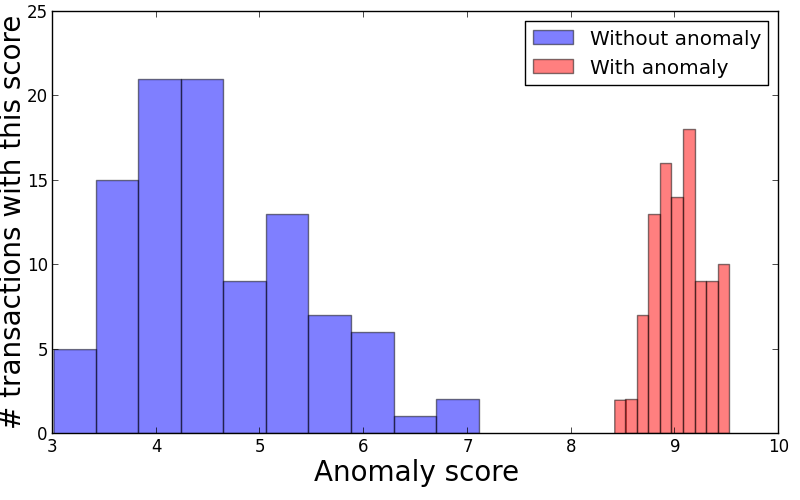}				
				\caption{\textbf{Significance of \scoreI scores.} The plot shows a clear separation between the highest \scoreI scores for random synthetic datasets with and without class 2 anomalies.}
				\label{fig:distr}
		\end{figure}

		In Figure~\ref{fig:distr} we show the distribution of the highest scores for both the datasets with and without an anomaly and with pattern supports in range [7-14\%] and an anomaly generator support at 28\%. We can see a clear distinction between the scores for `normal' and anomalous transactions.

	\subsection{Real World Data}
		To show that class 2 anomalies actually exist, are not identified by the state-of-the-art in anomaly detection, and can give much insight we performed multiple experiments on real world datasets from various domains. We used the \textit{Adult}, \textit{Zoo} and \textit{Bike Sharing} datasets from the UCI repository,\!\footnote{\url{http://archive.ics.uci.edu/ml/datasets.html}} together with the \textit{Mammals} \cite{mitchell-jones:societas} en \textit{ICDM Abstracts} \cite{debie:11:dami} datasets.
	
		\subsubsection*{Adult}
			The \textit{Adult} dataset contains information about 48\,842 people about their age, education, occupation, marital-status and more and is used to predict whether someone's income exceeds \$50K a year. 
			
			We computed a ranking based on the \scoreI score and found some interesting anomalies. The top-ranked transaction contains the very unexpected co-occurrence of someone for which the attribute sex is female yet for whom the relationship status has the value of husband. The following 3 anomalies are persons with a similar situation but with the patterns reversed. That is, the dataset contains 3 persons who's sex is male and who's relationship is wife. The \scoreS rankings of these first 4 people are 115, 148, 89 and 4\,090, respectively. These examples show that class-2 anomalies indeed exist in real datasets, and that \scoreI is effective at identifying these -- whether for further investigation, or data cleaning. 			
			
\begin{figure}[t]
				\centering
					\includegraphics[width=8cm]{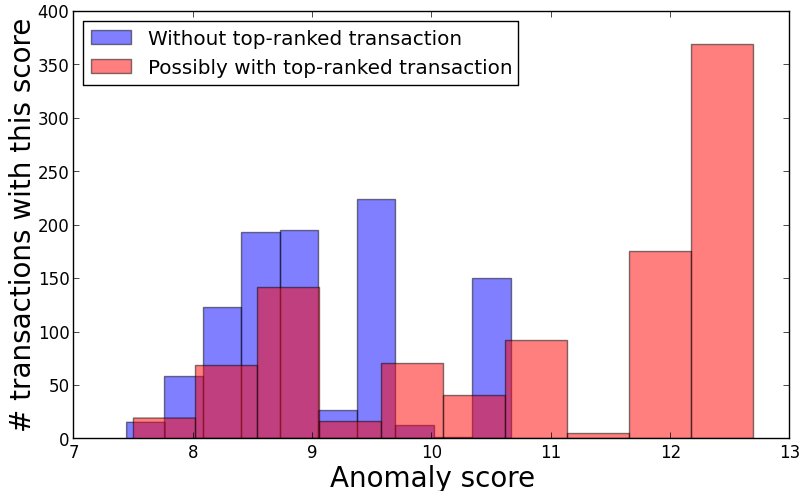}				
					\caption{\textbf{Significance test on \textit{Adult} dataset.} This plot shows the difference in the distribution of highest scores for bootstrap samples without (blue) and possibly with (red) the highest ranked transaction from the original dataset.}
					\label{fig:sig-adult}
			\end{figure}

			To get an idea of the significance of these results we performed the significance test as described in Section~\ref{sec:sig}. Figure~\ref{fig:sig-adult} shows the difference in the distribution of highest scores for bootstrap samples without (blue) and possibly with (red) the top-ranked transaction from the original dataset. Figure~\ref{fig:sig-adult} gives insight in how much this transaction deviates from the norm, as the difference between the two distributions can only be caused by this transaction.
					
		\subsubsection*{Zoo}
			The \textit{Zoo} data contains 17 attributes describing 101 different animals. 
			
\begin{figure*}[h]
				\centering
					\includegraphics[width=17cm]{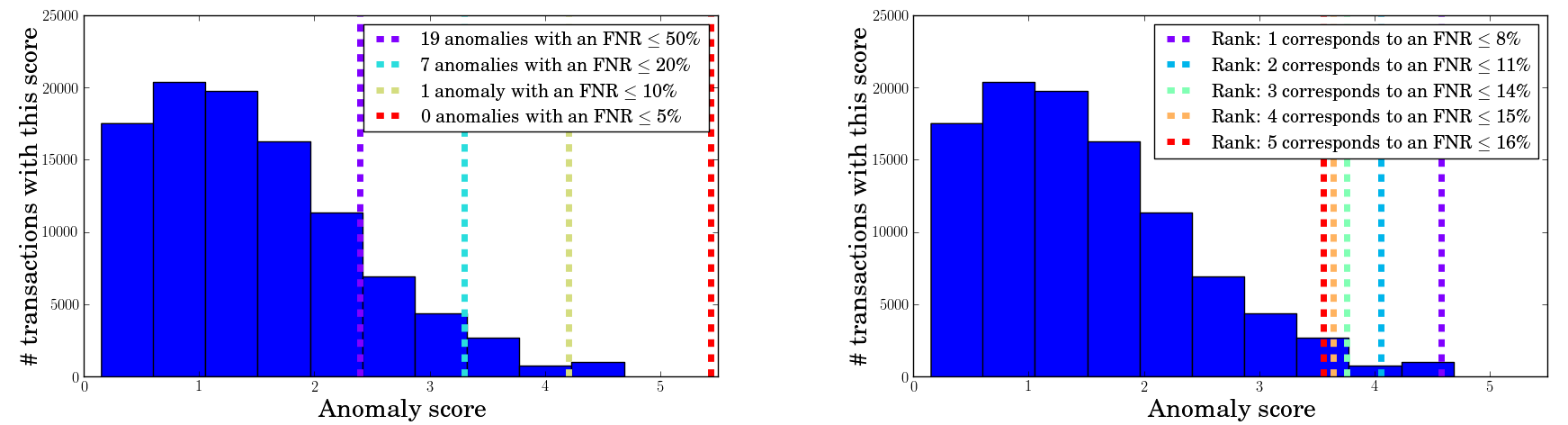}
					\caption{\textbf{Anomalies in the \textit{Zoo} dataset.} The histogram shows the estimated distribution of anomaly scores. (left) The vertical lines show the decision thresholds at false-negative rates of 50\%, 20\%, 10\% and 5\%, together with the number of original transactions that score above the threshold. (right) The vertical lines show the scores of the top-5 ranked anomalies in the original dataset, together with the false-negative rates corresponding to the decision threshold for their score.}					
					\label{fig:thresh}
			\end{figure*}

			We performed the bootstrap method described in Section~\ref{sec:inves} to determine which transactions are worth investigating. To this end, we generated 1\,000 bootstrap samples for which we computed the anomaly scores for all transactions. In Figure~\ref{fig:thresh} we show the distribution of all these scores with a histogram. Further, in the left plot we show the threshold ($\theta$) values corresponding to false-negative rates (FNR) of 50\%, 20\%, 10\% and 5\% respectively, together with the number of transactions from the original dataset that score above $\theta$. In the right plot we show the anomaly scores of the 5 highest ranked transactions in the original dataset together with the FNR corresponding to a $\theta$ equal to their score.
			In Figure~\ref{fig:thresh} in the left plot we see that with an FNR below 10\% only the top-ranked transaction scores above $\theta$. This transaction contains information about the platypus (duck bill) and from our results we found that the co-occurrence causing this high score is that the platypus is the only oviparous mammal in the dataset. In Figure~\ref{fig:thresh} in the right plot we see that the chance that the second ranked animal belongs to the positive class is less than 11\%. This is the scorpion for which \scoreI found that it is the only animal without teeth that is not oviparous. 			 

		\subsubsection*{ICDM Abstracts}
			Next we ran our algorithm on the \textit{ICDM Abstracts} dataset. This dataset consists of the abstracts from the ICDM conference, after stemming, and removing stopwords.\!\footnote{The data is available upon request from the author of~\cite{debie:11:dami}.}
			
			For this data we would expect co-occurrences of terms used in different research (sub)fields to rank highly. In Table~\ref{tab:icdm} we show the top 5 highest ranked abstracts with their explanation. That is, we show the unexpected co-occurrence responsible for the high \scoreI score. Further, only to show that these class 2 anomalies are not identified by the state-of-of-the-art, we show their \scoreS rank. The abstract with the highest \scoreI rank contains both the frequently used words `pattern mining' and `training', which is an unexpected combination. After reading the corresponding abstract it appears that the term `training' was used to refer to physical exercise rather than that of an algorithm. Other highly ranked abstracts show similar unexpected co-occurrences, for example `learning' on one side and `frequent pattern mining' on the other or `frequent pattern mining' and `compare', which suggest that exploratory algorithms are difficult to compare.
					
			\begin{table*}[h] 		
				\centering
				\caption{\textbf{The top 5 out of 859 abstracts from \textit{ICDM Abstracts}, with the corresponding unexpected co-occurrences explaining the high \scoreI scores.} Next to the \scoreI rank also the \scoreS rank is reported to show that class 2 anomalies are not identified, but ranked low, using an algorithm constructed for class 1 anomalies. }
				\label{tab:icdm}				
				\begin{tabular}{cccccc} 
					\toprule
					\textbf{\scoreI} &&&& \textbf{\scoreS} \\ 
					\cmidrule{1-3}
					\cmidrule{5-5}					
					& \multicolumn{2}{l}{\textbf{Most unexpected co-occurrence explaining the anomaly score}} && \\
					\cmidrule{2-3}	
					Rank & Pattern A & Pattern B && Rank 	 \\
					\midrule	
					1	&	['mine', 'pattern']	&	['train'] && 165	\\
					2	&	['algorithm', 'mine', 'pattern', 'frequent'] &	['learn'] && 183 \\
					3	&	['rule'] &	['vector'] && 132	\\
					4	&	['frequent', 'itemset'] &	['learn'] && 193	\\
					5	&	['mine', 'pattern', 'frequent'] &	['compar'] && 556	\\
					\bottomrule		
				\end{tabular}	
			\end{table*}

		\subsubsection*{Mammals}
			The \textit{Mammals} dataset consists of presence/absence records of 121 European mammals within 2\,183 geographical areas of 50 $\times$ 50 kilometres.\!\footnote{The full dataset~\cite{mitchell-jones:99:atlas} is available upon request from the Societas Europea Mammalogica, \url{http://www.european-mammals.org}.} In this dataset an anomaly constitutes two large territories of (groups of) animals which only overlap in a small region. 
			
\begin{figure}[h]
				\centering
					\includegraphics[width=8cm]{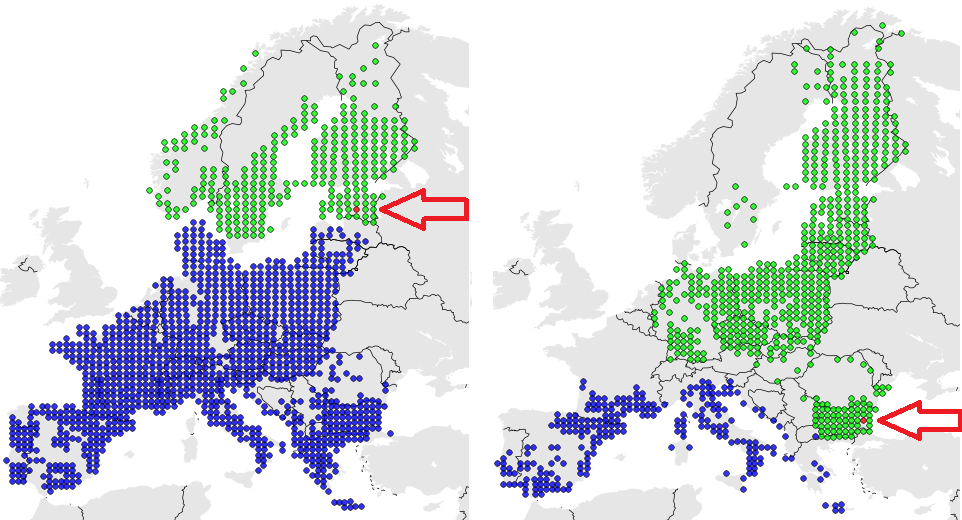}				
					\caption{\textbf{\scoreI in action; top-ranked anomalies on the \textit{Mammals} dataset.} The explanations for these highly ranked area's are as follows. On the left we see that the habitat of the beech marten (blue) only intersects with that of the moose, European hedgehog and mountain hare (green) at the (red) area pointed to by the arrow. On the right we see the habitat of the Etruscan shrew (blue) only intersects with that of the raccoon dog (green) at the (red) area pointed to by the arrow.}
					\label{fig:mam}
			\end{figure}

			Figure~\ref{fig:mam} shows two top-ranked area's (in red and pointed to by arrows) and readily explains why these are anomalous. For each of these two area's two groups of animals share this territory where the rest of their territory is completely separated. On the left in Figure~\ref{fig:mam} we see that the large habitat of the beech marten intersects with that of the moose, the European hedgehog and the mountain hare only in this single area. On the right in Figure~\ref{fig:mam} we see a similar phenomenon for the Etruscan shrew on one side and the raccoon dog on the other. The ranks of these two areas using \scoreS are 591 and 294 out of the 2\,183, respectively. There are also top-ranked areas that are explained by two groups of animals which habitat intersects in multiple areas (of course including the area that has received this score).
			
		\subsubsection*{Bike Sharing}
			The \textit{Bike Sharing} dataset contains the daily count of rental bikes in the years 2011 and 2012 in the Capital Bikeshare system with corresponding weather and seasonal information. 
			
			The \scoreI score of the highest ranked day in the dataset is not the result of any rental behaviour, but shows a very rare co-occurrence of a relatively low real temperature in combination with a relatively high apparent (perceived) temperature. This indeed seems strange as people more often feel colder as a result of wind-chill. Although this anomaly gives us no information about bike sharing, it is an actual class 2 anomaly present in the data.

\section{Discussion} \label{sec:dis}
	The experiments show that although the state-of-the-art in anomaly detection is not able to identify the newly defined class 2 anomalies, we can identify them using our new \scoreI score. We demonstrated that a naive baseline approach using closed frequent items as input set quickly becomes infeasible when the number of patterns present in the data grows. Using a \Slim pattern set to compute our \scoreI score, however, we attain similar results in a fraction of the time. We showed the statistical power of our method which scores transactions containing planted class 2 anomalies significantly higher than `normal' transactions. Moreover, both on transaction and categorical synthetic data we showed that \scoreI always ranked the planted anomaly at the top.
	
	From our experiments on real world datasets we find that the class 2 anomalies do actually exist and can provide useful insights. That is, because next to identifying interesting transactions the \scoreI score also readily explains which co-occurrence of patterns is responsible for the transaction's anomaly score. For example, in the \textit{Adult} dataset we found a very unexpected individual who is described as being a female husband. Further we showed how a \scoreI ranking can be used to study the significance of identified anomalies using a bootstrap approach. For example, in the \textit{Zoo} dataset we found that the platypus, which is special because its the only oviparous mammal, has a less than 8\% chance on being `normal' given the data. Each of these class 2 anomalies were not identified, i.e.\ ranked low, using \scoreS or \Comprex.

\section{Conclusion} \label{sec:con}
	In this paper we introduced a new class of anomalies which we refer to as unexpected co-occurrences of patterns. We showed that the anomalies identified by state-of-the-art in anomaly detection are of a different class and that these methods are not able to identify unexpected co-occurrences of patterns. We introduced the \scoreI score which intuitively scores a transactions based on its most unexpected co-occurrence of patterns. Using \scoreI we ably identify all planted anomalies in synthetic data and find interesting explanations for anomalous transactions in real world data. Besides useful for  identifying interesting behaviour, \scoreI also makes it possible to detect errors in data that previous methods cannot, making it also very suited for data cleaning purposes.

\section*{Acknowledgments}
	Roel Bertens and Arno Siebes are supported by the Dutch national program COMMIT. 
	Jilles Vreeken is supported by the Cluster of Excellence ``Multimodal Computing and Interaction'' within the Excellence Initiative of the German Federal Government.

\balance

\bibliographystyle{abbrv}

\bibliography{abbreviations,references,bib-jilles}

\end{document}